\pdfoutput=1

\PassOptionsToPackage{table,dvipsnames}{xcolor}

\documentclass[11pt]{article}

\usepackage[final]{acl}
\usepackage{xspace} %
\usepackage{times}
\usepackage{latexsym}
\usepackage[T1]{fontenc}
\usepackage[utf8]{inputenc}
\usepackage{longtable}
\usepackage{microtype}

\usepackage{hyperref}
\usepackage{booktabs}
\usepackage{graphicx}
\graphicspath{{./imgs/}}
\usepackage{footmisc}
\usepackage{microtype}
\usepackage{amsmath}
\usepackage{amsfonts,bm}
\usepackage{xspace}

\newcommand{\Ni}{({\em i})~}
\newcommand{\Nii}{({\em ii})~}
\newcommand{\Niii}{({\em iii})~}
\newcommand{\Niv}{({\em iv})~}
\newcommand{\Nv}{({\em v})~}

\definecolor{mypink3}{cmyk}{0, 0.7808, 0.4429, 0.1412}

\makeatletter   
\newcommand{\sveryshortarrow}[1][3pt]{\mathrel{%
    \vcenter{\hbox{\rule[-.5\fontdimen8\scriptfont3]
               {\scriptratio\dimexpr#1\relax}{\fontdimen8\scriptfont3}}}%
   \mkern-4mu\hbox{\let\f@size\sf@size\usefont{U}{lasy}{m}{n}\symbol{41}}}}
\makeatother

\def\eqref#1{equation~\ref{#1}}

\def\1{\bm{1}}

\def\m1{{\bm{1}}}

\DeclareMathAlphabet{\mathsfit}{\encodingdefault}{\sfdefault}{m}{sl}
\SetMathAlphabet{\mathsfit}{bold}{\encodingdefault}{\sfdefault}{bx}{n}

\def\gD{{\mathcal{D}}}

\usepackage{arydshln}
\usepackage{subcaption}
\usepackage{caption}
\usepackage{array, makecell} %
\usepackage{footmisc}
\usepackage{alltt}
\usepackage{floatrow}
\usepackage{pifont}%
\usepackage{tabularx}
\usepackage{adjustbox}
\usepackage{multirow}
\usepackage{multirow, colortbl, caption}
\definecolor{lightblue}{RGB}{232, 244, 248}
\definecolor{lightpink}{RGB}{254, 238, 237}

\definecolor{bluelink}{RGB}{0,113,188}
\definecolor{greenlink}{RGB}{0,188,113}

\usepackage{tikz}

\usepackage{collcell}

\usepackage{etoolbox}

\newtoggle{inTableHeader}%
\toggletrue{inTableHeader}%
\newcommand*{\StartTableHeader}{\global\toggletrue{inTableHeader}}%
\let\OldTabular\tabular%
\let\OldEndTabular\endtabular%
\renewenvironment{tabular}{\StartTableHeader\OldTabular}{\OldEndTabular\StartTableHeader}%

\newcommand*{\MinNumber}{-1.0}%
\newcommand*{\MidNumber}{0.0} %
\newcommand*{\MaxNumber}{1.0}%

\newcommand{\ApplyGradient}[1]{%
  \iftoggle{inTableHeader}{#1}{
    \ifdim #1 pt > \MidNumber pt
        \pgfmathsetmacro{\PercentColor}{max(min(100.0*(#1 - \MidNumber)/(\MaxNumber-\MidNumber),100.0),0.00)} %
        \hspace{-0.33em}\colorbox{yellow!\PercentColor!blue}{#1}
    \else
        \pgfmathsetmacro{\PercentColor}{max(min(100.0*(\MidNumber - #1)/(\MidNumber-\MinNumber),100.0),0.00)} %
        \hspace{-0.33em}\colorbox{blue!\PercentColor!blue}{#1}
    \fi
  }}
\newcolumntype{R}{>{\collectcell\ApplyGradient}c<{\endcollectcell}}

\usepackage[nameinlink]{cleveref}
\crefformat{section}{\S#2#1#3} %
\crefname{algorithm}{Alg.}{Algs.}
\crefname{table}{Table}{Tables}
\crefformat{subsection}{\S#2#1#3}
\Crefname{equation}{Eq.}{Eqs.}
\Crefname{figure}{Figure}{Figures}

\usepackage[colorinlistoftodos,prependcaption,textsize=tiny]{todonotes}

\usepackage{soul}

\usepackage{float}

\usepackage{multirow}%
\usepackage{hhline}%

\usepackage{amssymb}   %
\usepackage{stmaryrd}  %

\definecolor{headerLavender}{RGB}{230, 230, 250} %
\definecolor{rowLightGray}{RGB}{245, 245, 245} %
\definecolor{rowCream}{RGB}{255, 250, 240} %
\definecolor{errorRed}{RGB}{255, 77, 77} %

\definecolor{chartqapro1}{RGB}{30,160,220} %
\definecolor{chartqapro2}{RGB}{50,200,100} %

\definecolor{dashqa1}{RGB}{63,81,181}   %
\definecolor{dashqa2}{RGB}{0,188,212}   %

\newcommand{\dashInteractqa}[1]{\textsc{
\textcolor{dashqa1}{Dashboard}\textcolor{dashqa2}{QA}}}

\title{
\dashInteractqa{}: Benchmarking Multimodal Agents for Question Answering on Interactive Dashboards
}

\author{
Aaryaman Kartha$^{\clubsuit}$ \thanks{\ \ Equal contribution.}, Ahmed Masry$^{\clubsuit}$ \footnotemark[1], Mohammed Saidul Islam$^{\clubsuit}$, 
Thinh Lang$^{\clubsuit}$,  \\ \bf Shadikur Rahman$^{\clubsuit}$, Ridwan Mahbub$^{\clubsuit}$, Mizanur Rahman$^{\clubsuit\bigstar}$, Mahir Ahmed$^{\clubsuit}$, \\ \bf Md Rizwan Parvez$^{\S}$, Enamul Hoque$^{\clubsuit}$, Shafiq Joty$^{\diamondsuit\triangle}$ \\
$^\clubsuit$York University, Canada, 
$^\bigstar$RBC, Canada,
$^\S$Qatar Computing Research Institute (QCRI)  \\
$^\diamondsuit$Nanyang Technological University, Singapore, 
$^\triangle$Salesforce Research, USA \\
\{aarykary, masry20, saidulis, tlang46, shadikur, mrahmed, rmahbub, enamulh\}@yorku.ca \\
\{mparvez@hbku.edu.qa, sjoty@salesforce.com\}
}

\begin{document}

\maketitle

\begin{abstract} 

Dashboards are powerful visualization tools for data-driven decision-making, integrating multiple interactive views that allow users to explore, filter, and navigate data. 
Unlike static charts, dashboards support rich interactivity, which is essential for uncovering insights in real-world analytical workflows. However, existing question-answering benchmarks for data visualizations largely overlook this interactivity, focusing instead on static charts. This limitation severely constrains their ability to evaluate the capabilities of modern multimodal agents designed for GUI-based reasoning.
To address this gap, we introduce \dashInteractqa{}, the first benchmark explicitly designed to assess how vision-language GUI agents comprehend and interact with real-world dashboards. 
The benchmark includes %
405 question-answer pairs with interactive dashboards spanning five categories: multiple-choice, factoid, hypothetical, multi-dashboard, and conversational. By assessing a variety of leading closed- and open-source GUI agents, our analysis reveals their key limitations, particularly in grounding dashboard elements, planning interaction trajectories, and performing reasoning. Our findings indicate that interactive dashboard reasoning is a challenging task overall for all the VLMs evaluated. Even the top-performing agents struggle; for instance, the best agent based on Gemini-Pro-2.5 achieves only 38.69\% accuracy, while the OpenAI CUA agent reaches just 22.69\%, demonstrating the benchmark's significant difficulty. We release \dashInteractqa{} at \href{https://github.com/vis-nlp/DashboardQA}{https://github.com/vis-nlp/DashboardQA}
. %

\end{abstract}

\section{Introduction}

 \begin{figure*}[t!]
    \includegraphics[width=\textwidth]{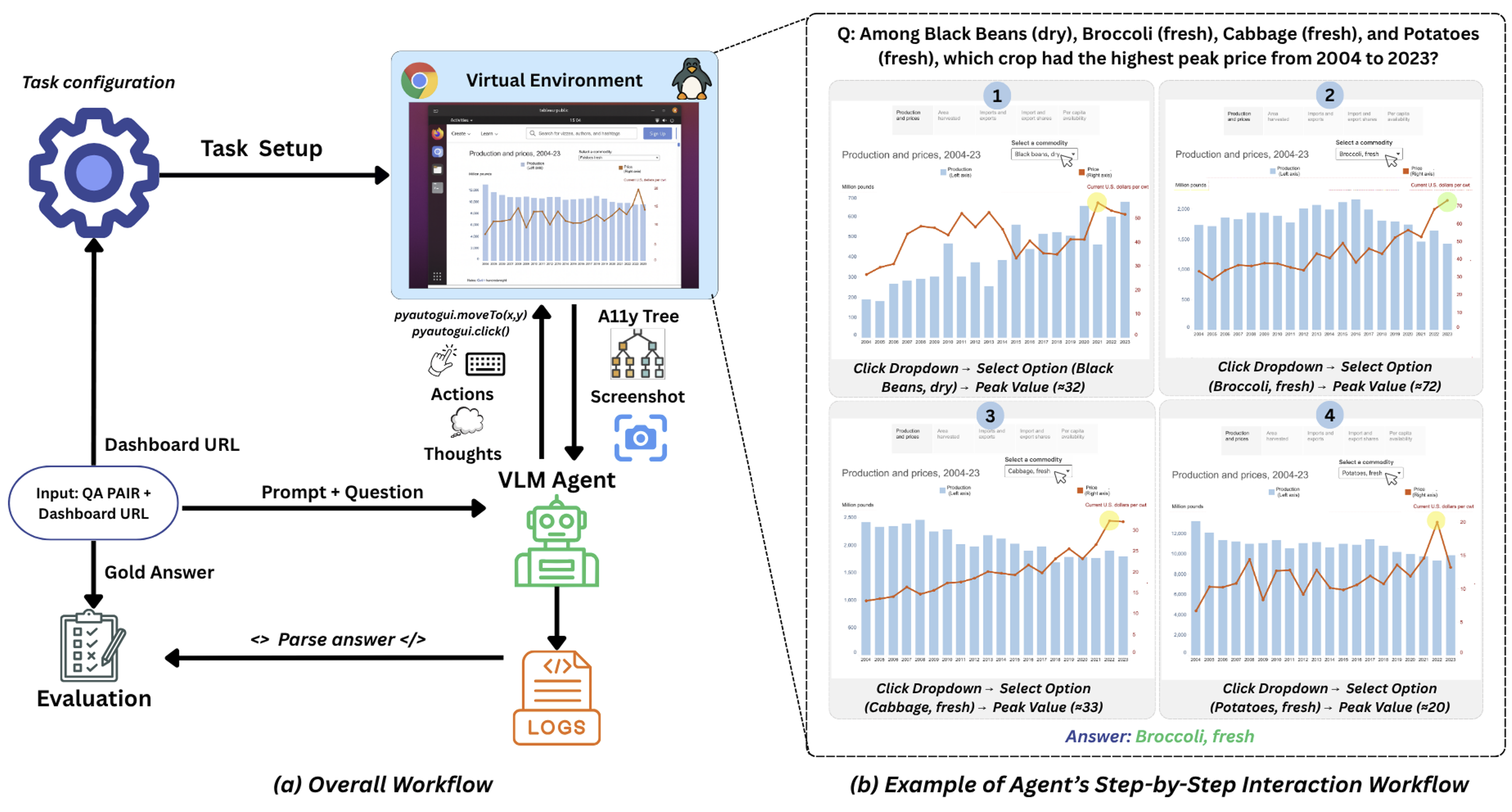}
    \caption{An overview of the \dashInteractqa~ task execution pipeline within the OSWorld environment, where (a) illustrates the overall setup of the VLM agent, and (b) shows an example interaction trajectory required to find an answer to a given question. The VLM, acting within an Ubuntu virtual environment, takes visual (Screenshot) or structural (A11y tree) input at each step and guides the navigation actions by generating corresponding \texttt{pyautogui} commands at every step. The action space of our agent consists of moving the mouse to a particular location, scrolling, and clicking.
    }
    \label{fig:examle-task}
\end{figure*}

Using visualizations to explore data and answer questions is fundamental to many decision-making tasks \citep{10.1145/3313831.3376467}. While static charts effectively present snapshots of specific insights, they fall short in capturing the dynamic, iterative nature of real-world data exploration. In practice, analysts continuously refine their queries, switch views, and filter subsets of data as their investigative goals evolve \citep{8019833}. {Interactive dashboards} support these workflows by enabling users to adjust visual and query parameters via UI controls (e.g., filters, sliders, dropdowns, tabs), while linking multiple coordinated views such that interactions in one view propagate to others—facilitating navigation through complex datasets and extraction of deeper insights.
However, question answering over such dashboards presents unique challenges. Answering a single question may require a sequence of GUI operations such as setting filters, switching tabs, zooming into a chart, then computing a ratio or identifying an outlier across views, while maintaining and interpreting intermediate states (see an example in Figure \ref{fig:examle-task}). This multi-step process is cognitively demanding for humans, incurring high mental load as they make efforts to manage linked visualizations and interactive constraints~\citep{convertino2003exploring}. These challenges raise a fundamental research question: \textit{Can we build AI agents that answer questions by interacting with dashboards as humans do—through a sequence of informed, goal-directed GUI operations over dynamic visual states?} 

While interest in benchmarking reasoning over data visualizations has grown over the past few years, existing work has consistently overlooked the critical element of interactivity. Early benchmarks focused on synthetic static charts (FigureQA \cite{kahou2017figureqa}, PlotQA \cite{plotqa}), and while later work introduced real-world data and more complex reasoning (ChartQA \cite{masry2022chartqa}, ChartQAPro \cite{masry2025chartqaprodiversechallengingbenchmark}), the visualizations remained static. In parallel, while vision-language models (VLMs) have significantly advanced static chart understanding \cite{openai2024gpt4technicalreport, geminipro25,masry2024chartgemma}, their application to interactive data analysis remains largely unexplored.

Concurrently, a wave of general-purpose GUI benchmarks like OSWorld~\citep{xie2024osworld}, AndroidWorld~\citep{rawles2025androidworlddynamicbenchmarkingenvironment}, and OfficeBench~\citep{officebench} has emerged, revealing that even state-of-the-art agents struggle with fundamental skills like grounding UI elements and planning long interaction sequences.

This has spurred the development of a new class of agentic approaches, from closed-source systems like OpenAI's Computer-Use Agent \citep{openaiComputerUsingAgent} to open-source agents like UI-TARS \citep{uitars} and hybrid strategies pairing powerful planners (e.g., Gemini Pro 2.5~\citep{geminipro25} or GPT-4o~\citep{openai2024gpt4technicalreport}) with specialized visual grounding modules like JEDI~\citep{jedi} or GTA1~\citep{gta1}. Yet, existing GUI benchmarks to evaluate these agents remain focused on utility-oriented tasks (e.g., clearing cookies, editing files), offering limited insights into agents' ability to perform analytical reasoning over visual data.

To bridge this gap, while also promoting the development of more advanced GUI agents, we introduce \dashInteractqa{}, the first benchmark for agentic question answering over interactive dashboards. It comprises 112 diverse dashboards from Tableau Public and 405 expert-authored questions including factoid, multiple-choice, hypothetical, conversational, and unanswerable questions to reflect realistic, complex analytical workflows. Answering these questions often requires multi-step interaction sequences such as filtering, switching views, interpreting updated charts, and analyzing various data visualization types present. We evaluate a range of open- and closed-source agents on this benchmark and find that even top-performing models struggle significantly; for example, the best-performing agent based on Gemini-Pro-2.5 achieves only 38.69\% accuracy, underscoring the benchmark’s difficulty and novelty.

In summary, our contributions are:
\Ni \textbf{A comprehensive benchmark} that requires complex, multi-step GUI interactions and more advanced visual reasoning with coordinated multi-view dashboards to answer questions.
\Nii \textbf{Extensive evaluation} of open- and closed-source agents, revealing significant performance drops on interactive dashboard QA tasks compared to static charts and utility-based benchmarks;
\Niii \textbf{In-depth analysis and ablations} that reveal key limitations in grounding, planning, and visual reasoning, providing concrete directions for improving future multimodal agents.

\vspace{-0.5em}
\section{Related Work}
\vspace{-0.5em}
\begin{figure*}[t!]
    \centering
    \includegraphics[width=\textwidth]{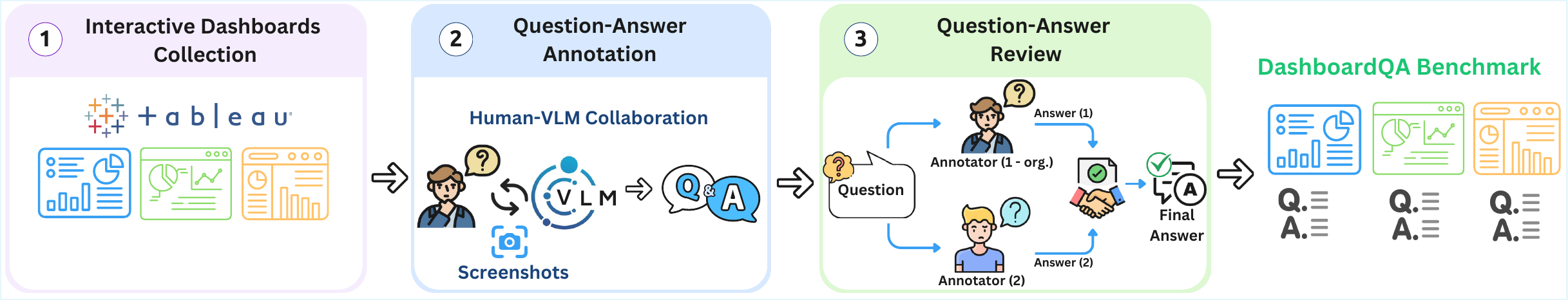}
    \caption{\dashInteractqa{} benchmark construction process. We collect interactive dashboards from the Tableau Public website. Annotators collaborate with VLMs to generate question–answer pairs by providing dashboard screenshots. Each QA pair is then reviewed by a second annotator to verify and resolve the final answer.}
    \label{fig:data_const}
\end{figure*}

\paragraph{Data Visualization QA Benchmarks.}

Benchmarks designed to evaluate visual-language models (VLMs) through question answering on data visualizations have significantly evolved in complexity over recent years. Early datasets, such as FigureQA~\cite{kahou2018figureqaannotatedfiguredataset}, DVQA~\cite{dvqa}, and PlotQA~\cite{plotqa}, primarily relied on synthetic data and simple charts. Later, benchmarks including ChartQA~\cite{masry2022chartqa} and InfoVQA~\cite{infovqa} introduced real-world visualizations paired with human-authored questions. However, rapid advancements in vision-language modeling quickly saturated performance on these datasets. To provide a greater challenge, recent benchmarks like CharXiv~\cite{wang2024charxivchartinggapsrealistic} and ChartQAPro~\cite{masry2025chartqaprodiversechallengingbenchmark} were developed, emphasizing more complex and diverse question types as well as richer visual content, including infographics and static dashboards.

Despite these advancements, existing benchmarks are fundamentally limited by their reliance on static images. Even recent work like Multi-ChartQA~\cite{zhu2025multichartqabenchmarkingvisionlanguagemodels}, which incorporates multiple charts, still operates in a non-interactive setting. This stands in contrast to real-world analysis, which is often performed on interactive settings featuring coordinated views, filters, and other dynamic controls. \dashInteractqa{} aims to bridge this critical gap by evaluating question answering on interactive dashboards, encompassing diverse question types and dynamic user interactions.

\paragraph{GUI Agents and Associated Benchmarks.}
Recent benchmarks such as OSWorld~\cite{xie2024osworldbenchmarkingmultimodalagents}, AndroidWorld~\cite{rawles2025androidworlddynamicbenchmarkingenvironment}, WebArena~\cite{zhou2024webarenarealisticwebenvironment}, and OfficeBench~\cite{officebench} evaluate agent capabilities in grounding (locating and manipulating UI elements) and planning (sequencing actions) across desktop, mobile, web, and office environments. Specialized grounding benchmarks such as ScreenSpot and ScreenSpot Pro further isolate the grounding challenge~\cite{screenspot, screenspotpro}.

Existing GUI agent approaches typically either use separate specialized models for grounding and planning or unified models for both tasks.  Closed-source models are generally strong in planning but may struggle with grounding, leading to the development of open-source models explicitly trained for grounding, such as JEDI, AgentS2, and GTA1 \cite{jedi, agents2, gta1}. Conversely, unified models include fully open-source solutions like UI-TARS \cite{uitars} and proprietary models such as CUA, Gemini, GPT-4o, and Claude \cite{openaiComputerUsingAgent, geminiteam2024gemini15unlockingmultimodal, openai2024gpt4technicalreport, claude}

Existing benchmarks do not evaluate the end-to-end analytical reasoning required for interactive dashboards—skills like grounding linked views, maintaining state across interactions, and performing visual mathematical calculations.\dashInteractqa{} is designed specifically to test GUI agents and VLMs on these critical capabilities.

\section{\textsc{The  \dashInteractqa{} Benchmark}}

As illustrated in Figure~\ref{fig:examle-task}, an agent is given a dashboard interface and a natural language question about it. To arrive at the correct answer, the agent must reason and plan a sequence of actions--navigate the interface by making selections and applying filters, extract relevant information from the views, synthesizing information across views and perform any necessary calculations.

\subsection{Dataset Construction}

The dataset construction process comprises three key stages (\Cref{fig:data_const}): \Ni Interactive Dashboard curation, \Nii Question-Answer pair collection, and \Niii Question-Answer Review. 
\
\paragraph{Phase 1 - Dashboard Curation:}
Tableau public was chosen as the main source of our dashboards ~\cite{tableauPublic}, given the vast range of dashboards available to access. Tableau also supports rich interactivity features such as coordinated views, cross-filtering, and cross-highlighting, making it especially suitable for evaluating agents that must reason over dynamic visual updates.

We curated dashboards with multiple linked views and meaningful text-chart integration to reflect the exploratory workflows our benchmark targets.
Dashboards were collected using search terms such as “interactive dashboards,” “data dashboards,” and “analytics dashboards,” or by searching relevant topics such as "health" or "renewable energy." They were initially screened manually for relevance and popularity. We excluded static or single-view dashboards and decorative infographics, retaining only those with rich interactivity (e.g., dropdowns, filters, tabs), pertinent modern day topics, varied chart types, and visually distinct layouts. At the end, we selected 112 high-quality dashboards for our benchmark.

\begin{figure*}[t!]
    \centering
    \includegraphics[width=\textwidth]{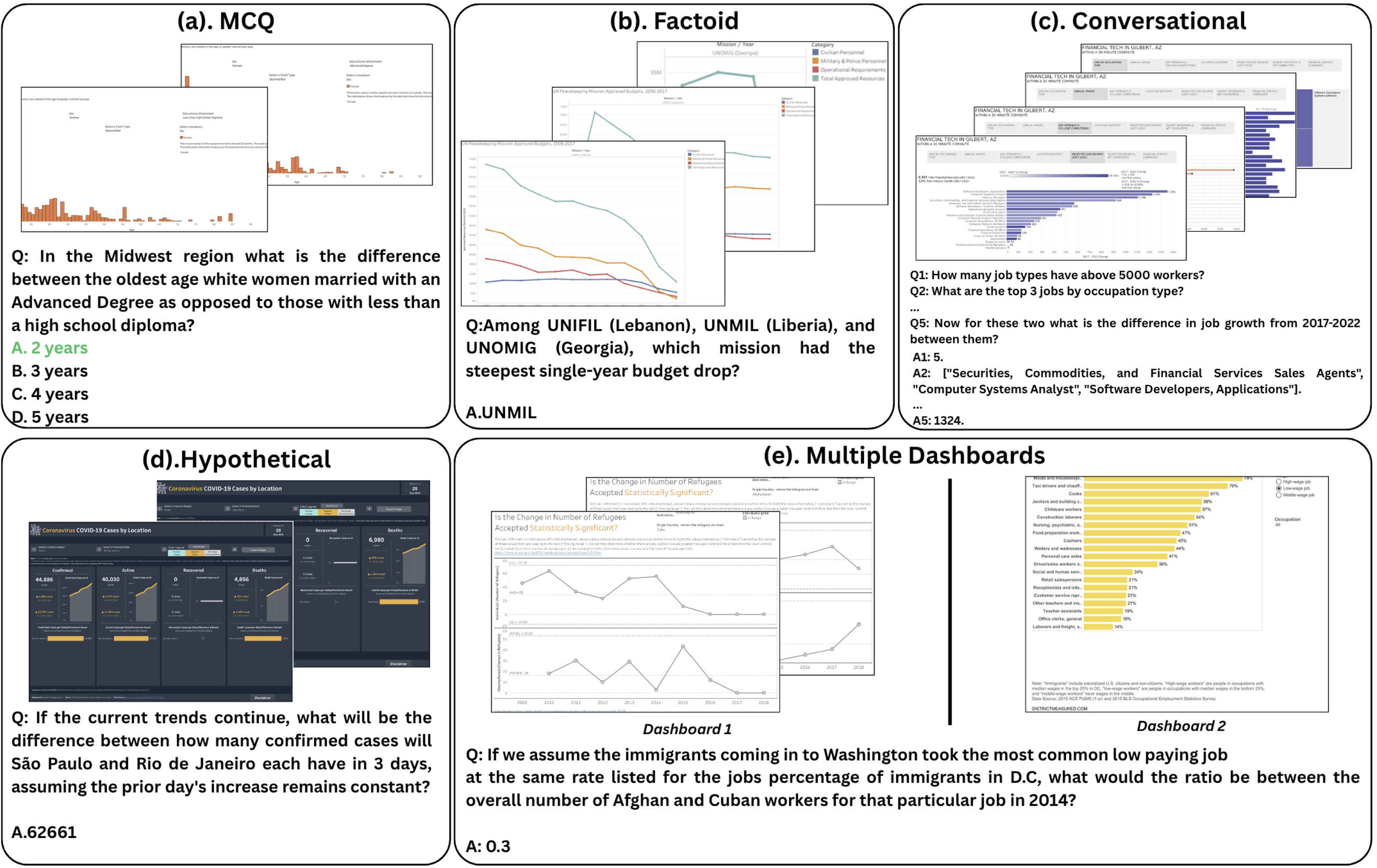}
    \caption{\label{fig:qa_types}DashboardQA focuses on 5 categories of questions that require exploration of dynamic dashboards. Reasoning between multiple views is required to derive the answer. 
    }
\end{figure*}

\paragraph{Phase 2 - QA pair creation:}
\dashInteractqa{} comprises five question categories: \Ni factoid, \Nii MCQ, \Niii Hypothetical, \Niv Conversational, and \Nv Multi-Dashboard (see examples in Figure  \ref{fig:qa_types}). 

A team of 8 annotators, comprised of university students and professors well versed in topics surrounding data visualization downstream tasks, collaboratively created these QA pairs, assigning two annotators per category to ensure semantic diversity. To enhance coverage and reduce bias, we adopted a human–VLM collaboration process comprising three steps:

$\bullet$ \,\textbf{Curating Seed QA Pairs:} Annotators crafted a diverse set of seed QA pairs covering different question types that required complex reasoning, interactive operations (filters, tabs), and realistic analytical workflows.\\
$\bullet$ \,\textbf{VLM-Assisted Expansion:} To enrich our dataset and reduce potential biases, we utilized multiple VLMs (GPT-4o, Gemini, Claude) to expand the initial seed set of QA pairs. Each model received a seed QA pair as input and was instructed to generate five additional QA pairs based on multiple dashboard views. This interactive prompting strategy encouraged greater diversity and novelty in the generated questions.\\
$\bullet$ \,\textbf{Human Refinement:} Finally, annotators reviewed the generated questions to remove incorrect or overly simple ones(e.g., direct data retrieval from charts) and revised those that were unclear, hallucinatory or ambiguous. 

During the QA creation process, we prioritized interactivity to distinguish our benchmark from traditional chart QA tasks \cite{masry2022chartqa}. Conventional chart QA datasets typically use static charts, allowing answers to be derived from a single visualization without interaction. In contrast, \dashInteractqa{} focuses on dynamic dashboards that require multi-step reasoning through navigation across linked views and the application of interactive controls (e.g., filters, tabs). This design evaluates not only answer correctness but also navigational efficiency, reflecting realistic analytical workflows and imposing higher cognitive demands.

Below we present a brief description of each question type:\\
\textbf{Factoid:}  These questions emphasize information retrieval, pattern recognition, and comparative reasoning, requiring the model to not only extract relevant facts but also identify underlying patterns and make informed inferences based on contextual cues.\\
\textbf{MCQ:} MCQs test a VLMs ability to evaluate various candidate options for a question. We specifically focus on scenarios where the VLM Agent is presented with four or five answer options and must synthesize information from multiple views within a dashboard in order to select the correct response. Often, options feature values very close to each other, so it would test the ability to extract data values from visualizations as accurately as possible when used in intermediate calculations for the answer.
\\
\textbf{MultiDashboard:} Multi-dashboard questions require VLM Agents to reason across two or more separate dashboards, introducing a significantly more complex cognitive challenge. This task goes beyond simply understanding the structure and content of an individual dashboard; it demands the ability to interpret, compare, and synthesize information across multiple visual and textual contexts. Such questions test the VLM's capacity for cross-dashboard navigation, multi-view integration, and higher-order reasoning, making them inherently more demanding than single-dashboard interactions.
\\
\textbf{Conversational:} Conversational question answering evaluates the capacity of VLM Agents to comprehend pragmatic and contextual nuances in dialogue. This task requires models to effectively leverage prior question-answer pair, including intermediate computations, to accurately interpret and respond to follow-up queries. Each conversational sequence comprises multiple interrelated QA pairs centered on a or multiple specific dashboard views, with each question logically building upon the preceding one. Such sequences enable a thorough assessment of the model’s ability to handle contextual dependencies, including coreference resolution, multi-step reasoning, and both logical and arithmetic inference.\\

\paragraph{Phase 3 - Question-Answer Review:}
To ensure the quality and accuracy of our QA pairs,  each QA pair was independently reviewed by an annotator from a different question category. Discrepancies were resolved through discussion, and unresolved cases were removed.  This cross-category assessment not only verifies factual correctness but also helps identify and eliminate any unnecessary complexity in question phrasing. Additionally, the states collected for each QA pair/task were examined to ensure the optimal trajectory (Number of states) for answer collection was used. For estimation-based questions, we consider a response acceptable if it falls within a margin of error of \textless 0.5\% from the reference answer. From a sample size of 367 qa pairs We had an inter annotator agreement rate of 74.93\%, meaning 24.86\%  had to go under revision again and modified.

\subsection{Dataset Analysis}

We analyze \dashInteractqa{} along two dimensions: dataset composition, covering the diversity of visualizations, topics, and question types, and interaction diversity, representing UI controls and the multi-step complexity of tasks.

\subsubsection{Dataset Composition}
\begin{figure}[t]
    \centering
    \includegraphics[width=0.9\columnwidth]{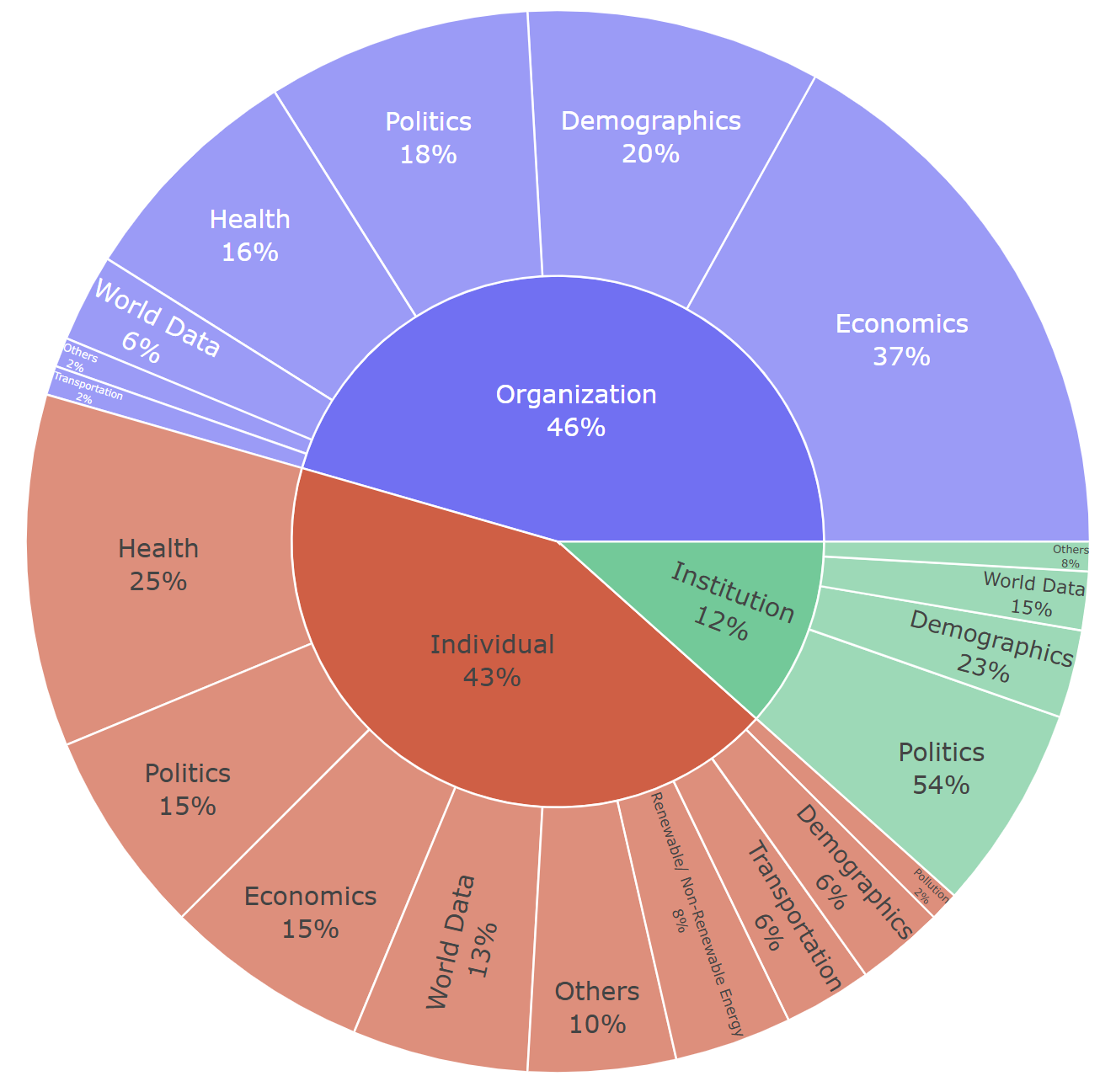}
     \caption{Dashboard topic breakdown by author type. For the 8 dashboard topics present, the nested pie chart shows the entire composition of the topics (Outer loop) to their respective author source type (Inner loop).
    }
    \label{fig:topic-diversity}
\end{figure}

The dataset dashboards were carefully curated to maximize high visual diversity, including 13 visualization types (Table ~\ref{tab:viz_table}). Fundamental chart types dominate: 48.51\% of dashboards include at least one line chart and 42.57\% include at least one bar chart. Other common charts in data science and analytics, such as maps, pie charts, scatter plots, and area charts, also appear with moderate frequency. The dataset   also exhibits strong domain diversity, covering a wide range of real-world topics and maintaining an almost equal distribution of dashboards created by organizations (Figure~\ref{fig:topic-diversity}). Finally, we also prioritized QA diversity, including five question categories: factoid, hypothetical, MCQ, multi-dashboard, and conversational. These categories are designed to test distinct reasoning skills, from direct retrieval to multi-step numerical and logical inference. 

\begin{table}[t]
\centering
\caption{Prevalence, count, and average number of respective visualization types across all dashboards, along with overall average of all visualizations.}
\label{tab:viz_table}
\resizebox{\columnwidth}{!}{%
\begin{tabular}{l|c|c|c}
\toprule
\textbf{Visualization} & \textbf{Prevalence} & \textbf{Count} & \makecell{\textbf{Avg. per} \\ \textbf{dashboard}} \\
\midrule
Area Chart     & 16.83\% & 31  & 0.28 \\
Bar Chart      & 42.57\% & 79  & 0.71 \\
Bubble Chart   & 5.94\%  & 7   & 0.06 \\
Bump Chart     & 0.99\%  & 1   & 0.01 \\
Donut Chart    & 1.98\%  & 2   & 0.02 \\
Heat Map       & 5.94\%  & 7   & 0.06 \\
Histogram      & 1.98\%  & 2   & 0.02 \\
Line Chart     & 48.51\% & 86  & 0.77 \\
Map            & 20.79\% & 25  & 0.22 \\
Pie Chart      & 6.93\%  & 9   & 0.08 \\
Scatter Plot   & 10.89\% & 30  & 0.27 \\
Crosstab       & 10.89\% & 12  & 0.11 \\
Tree Map       & 4.95\%  & 7   & 0.06 \\
\midrule
\textbf{Total} & \textbf{100\%} & \textbf{298} & \textbf{2.66} \\
\bottomrule
\end{tabular}
}
\end{table}

\subsubsection{ Interaction Diversity} 

We emphasize dynamic navigation, requiring agents to perform multi-step interactions rather than just single-view information extraction. To facilitate this, \dashInteractqa{} dashboards incorporate four primary UI controls: dropdown menus (92.31\%), tabs (23.08\%), radio buttons (7.69\%), and range sliders (7.69\%). Dropdown menus are the most prevalent, appearing in nearly all dashboards with an average of 2.38 each (Table~\ref{tab:navigation}).

\begin{table}[t!]
\centering
\caption{Prevalence, count, and average number of respective navigation tools across the dashboards curated.}
\label{tab:navigation}
\resizebox{\columnwidth}{!}{%
\begin{tabular}{l|c|c|c}
\toprule
\textbf{Navigation Tool} & \textbf{Prevalence} & \textbf{Count} & \makecell{\textbf{Avg. per} \\ \textbf{dashboard}} \\
\midrule
Dropdown Menus & 92.31\% & 266 & 2.38 \\
Radio Buttons  & 7.69\%  & 26  & 0.23 \\
Tabs           & 23.08\% & 103 & 0.92 \\
Range Sliders  & 7.69\%  & 9   & 0.08 \\
\bottomrule
\end{tabular}
}
\end{table}

\newcolumntype{C}{>{\centering\arraybackslash}X}

\begin{table}[t]
\centering
\caption{Distribution of views/states of dashboards required for the respective question categories.}
\label{tab:category_views}
\renewcommand{\arraystretch}{1.2}
\begin{tabularx}{\columnwidth}{l|*{5}{C}}
\toprule
\multirow{2}{*}{\textbf{Category Type}} & \multicolumn{5}{c}{\textbf{Views required to answer}} \\
\cmidrule(lr){2-6}
& \textbf{1} & \textbf{2} & \textbf{3} & \textbf{4} & \textbf{5+} \\
\midrule
MCQ             & 17 & 48 &  9 &  2 & 4 \\
Factoid         &  4 & 16 & 33 & 19 & 2 \\
Hypothetical    &  0 & 39 & 24 &  1 & 1 \\
Multi-Dashboard &  0 & 36 &  5 &  3 & 2 \\
Conversational  &  0 &  6 & 12 &  9 & 2 \\
\bottomrule
\end{tabularx}
\end{table}

In Table \ref{tab:category_views}, we show diversity based on task difficulty. In our setting, task difficulty is defined by the number of interactions—or "states"—an agent must navigate to find an answer, where each round of interactions (e.g., a dropdown selection followed by a radio button selection) creates a new dashboard view. Across categories, most questions require navigating through 2-3 states, with multi-dashboard questions demanding the most transitions, reflecting their need for deeper planning and cross-view reasoning.

\section{Methodology}
\subsection{Task Definition}
\label{subsec:task-definition}

\dashInteractqa{} frames the QA task as an interactive question-answering problem. Formally, given a dataset $\gD = {(d_i, q_i, a_i)}_{i=1}^{N}$, each example comprises an interactive dashboard $d_i$, an associated question $q_i$, and a ground truth verifibale answer $a_i$. The autonomous agent must navigate and interact with the dashboard $d_i$ to accurately generate the correct answer $a_i$. 

Successful completion of many of these tasks require significant planning, arithmetic and logical reasoning, as well as dynamic state tracking. Additionally, meta reasoning (Self-reflection) is needed within the reasoning trajectory for these dynamic, uncertain, dashboard environments.

\subsection{Overall Framework}
 We evaluate agents using an extended version of the OSWorld environment~\cite{xie2024osworld}, which facilitates interactive navigation within virtual machines. As illustrated in \Cref{fig:examle-task}, each evaluation begins by launching an Ubuntu OS virtual machine, opening the dashboard URL in a Chrome browser configured in full-screen mode at 1920×1080 resolution. An initial observation (state) of the environment is captured from the virtual environment and provided to the VLM agent, accompanied by a task instruction that includes an input prompt and the associated question from \dashInteractqa{}. The VLM agent processes this information and responds with predicted actions and intermediate reasoning ("thoughts"), which the environment then executes. Then, the cycles continue again by sending an observatin of the updated environemtn, ...etc. Below we define the observation and action spaces more formally.

\paragraph{Observation Space.}  
We employ two types of observations to represent the environment state to the VLM agent: \textit{(i)} \textbf{Screenshot}: a raw pixel-level capture of the dashboard interface, mimicking human perception, and \textit{(ii)} \textbf{Screenshot + Accessibility Tree (A11y)}: which augments the visual input with structured UI element information. This hybrid input is particularly beneficial for addressing grounding limitations commonly observed in VLMs when relying solely on raw visual data~\cite{xie2024osworld}. We extract the A11y tree using ATSPI\footnote{https://docs.gtk.org/atspi2/}. 

\paragraph{Action Space.}  
Following the OSWorld framework~\cite{xie2024osworld}, we define the action space to encompass a wide range of GUI operations, including mouse movements (e.g., clicking, dragging, scrolling) and keyboard inputs (e.g., typing text, pressing specific keys). Additionally, we incorporate three special meta-actions: \textbf{WAIT}, \textbf{FAIL}, and \textbf{DONE}. These allow the agent to \textit{(i)} pause and wait for the environment to update, \textit{(ii)} terminate the episode when the task is deemed unsolvable, or \textit{(iii)} signal task completion, respectively.

All responses from the agent are logged during this process. The process terminates when one of the following conditions is met: \emph{(i)}: the agent successfully completes the task and outputs the \textbf{DONE} meta-action, \emph{(ii)} the agent admits failure and outputs the \textbf{FAIL} meta-action, or \emph{(iii)} the agent reaches the max number of allowed steps, which is set to 25 steps.  

Finally, we extract the final answers from the VLM agent's logs (responses). The complete prompts used for models evaluation is presented in Appendix \ref{app:prompts-evaluation} and Table \ref{tab:prompt_templates}.

\section{Evaluation}

\subsection{Experimental Setup} 
\label{subsec:experimental-setup}

We conduct experiments under two observations types: dashboard screenshots (primary setting) and accessibility (a11y) trees using ATSPI\footnote{https://docs.gtk.org/atspi2/}. %
Evaluations are conducted on Google Cloud, deploying open-source models on either L4 or A100 GPU machines based on model size, utilizing the VLLM framework \citep{vllm}.

\subsection{Baselines}
We evaluate a comprehensive list of 12 agents categorized into three groups: closed-source, open-source, and hybrid. Closed-source models include GPT4o \cite{openai2024gpt4technicalreport}, O4-mini \cite{openai2024gpt4technicalreport}, Gemini Pro 2.5 ~\cite{geminipro25}, 
and OpenAI's Computer Use Agent (CUA) ~\cite{openaiComputerUsingAgent}. Hybrid models combine open-source grounding models with proprietary planning models and include Jedi-3B and 7B w/GPT4o \cite{jedi}.
Open-source models include UI-TARS 2B, and UI-TARS-1.5-7B \cite{uitars}.

\vspace{-1mm}
\subsection{Evaluation Metric} \label{subsec:evalmetric}
To evaluate model performance, we adopt the enhanced relaxed accuracy metric \cite{masry2025chartqaprodiversechallengingbenchmark}. This metric covers various answer formats: numerical responses allow a tolerance margin of 5\%, whereas answers referring to specific years require exact matches to prevent errors from minor differences (e.g., distinguishing between 1990 and 1993). For textual answers, the Average Normalized Levenshtein Similarity (ANLS) metric \cite{biten2019scenetextvisualquestion} is utilized.

\definecolor{human_baseline}{RGB}{255, 245, 170}
\definecolor{open_models_below_4B}{RGB}{185, 235, 255}
\definecolor{open_models_7B_12B}{RGB}{255, 219, 187}
\definecolor{closed_models}{RGB}{240, 240, 240}
\definecolor{chart_specific_models}{RGB}{217, 240, 211}

\begin{table*}[t]
\centering
\caption{Accuracy (\%) on \dashInteractqa{}  across different Observation Types (main headers) and Question Categories (sub-headers). Each Observation block includes five question types along with an Overall score. Color shading indicates model category: \colorbox{human_baseline!70}{human performance}, \colorbox{closed_models!70}{closed-source models}, \colorbox{open_models_below_4B!50}{hybrid models}, and \colorbox{open_models_7B_12B!50}{fully open-source models}. The highest score within each category is shown in bold.}
\resizebox{\textwidth}{!}{%
\begin{tabular}{l|cccccc|cccccc}
\toprule
\multirow{2}{*}{\textbf{Model}} 
& \multicolumn{6}{c|}{\textbf{Screenshot}} 
& \multicolumn{6}{c}{\textbf{Screenshot + A11y Accessibility Tree}} 
\\
\cmidrule(lr){2-7} \cmidrule(lr){8-13}
& \textbf{Factoid} & \textbf{MCQ} & \textbf{Convers.} & \textbf{Hypoth.} & \textbf{Multidash.} & \textbf{Overall}
& \textbf{Factoid} & \textbf{MCQ} & \textbf{Convers.} & \textbf{Hypoth} & \textbf{Multidash} & \textbf{Overall} \\
\midrule
\multicolumn{13}{l}{\textbf{\textit{Closed-Source Models}}} \\
\rowcolor{closed_models!50} GPT4-o & 8.88 & 20.00 & 14.81 & 7.69 & 4.35 & \cellcolor{closed_models}11.50 & 27.25 & 35.00 & 17.86 & 13.85 & 10.87 & \cellcolor{closed_models}22.94 \\
\rowcolor{closed_models!50} O4-mini & 1.35 & 1.25 & 0.00 & 0.00 & 0.00 & \cellcolor{closed_models}0.68 & 7.92 & 22.50 & 0.00 & 2.82 & 2.17 & \cellcolor{closed_models}9.14 \\
\rowcolor{closed_models!50} Gemini-Pro-2.5 & 10.33 & 15.00 & 22.22 & 7.69 & 8.70 & \cellcolor{closed_models}11.86 & 40.03 & 46.25 & 51.20 & 39.27 & 15.22 & \cellcolor{closed_models} 38.69 \\
\rowcolor{closed_models!50} OpenAI CUA & 26.45 & 42.50 & 29.63 & 7.22 & 0.00 & \cellcolor{closed_models} 22.69 & 22.24 & 38.75 & 16.93 & 3.08 & 0.00 & \cellcolor{closed_models}18.50 \\
\midrule
\multicolumn{13}{l}{\textbf{\textit{Open-Source Models}}} \\
\rowcolor{open_models_below_4B!50} Jedi-3B w/GPT4o & 32.70 & 40.00 & 44.85 & 23.41 & 6.67 & \cellcolor{open_models_below_4B}29.73& N/A & N/A & N/A & N/A & N/A  & \cellcolor{open_models_below_4B} N/A \\
\rowcolor{open_models_below_4B!50} Jedi-7B w/GPT4o & 35.33 & 42.50 & 49.52 & 34.00 & 2.17 & \cellcolor{open_models_below_4B}33.09 & N/A & N/A & N/A & N/A & N/A & \cellcolor{open_models_below_4B}N/A \\
\rowcolor{open_models_7B_12B!50} UI-Tars-2B & 0.75 & 0.00 & 0.00 & 0.00 & 0.00 & \cellcolor{open_models_7B_12B}0.19 & N/A & N/A & N/A & N/A & N/A & \cellcolor{open_models_7B_12B}N/A \\
\rowcolor{open_models_7B_12B!50} UI-Tars-1.5-7B & 8.98 & 12.50 & 0.00 & 2.40 & 0.00 & \cellcolor{open_models_7B_12B}6.23 & N/A & N/A & N/A & N/A & N/A & \cellcolor{open_models_7B_12B}N/A \\
\bottomrule
\end{tabular}
}
\label{tab:prompt-question-accuracy}
\end{table*}

\vspace{-1mm}
\subsection{Main Results}
Table~\ref{tab:prompt-question-accuracy} presents accuracy across two observation settings: Screenshots and Screenshots with Accessibility Tree (A11y), and five question categories: Factoid, MCQ, Conversational, Hypothetical, and Multi-Dashboard. With A11y, Gemini Pro 2.5 achieves the strongest overall accuracy (40.80\%), outperforming other closed-source agents (GPT-4o: 25.68\%, OpenAI CUA: 19.33\%). Under the Screenshots-only setting, all closed-source agents remain below 21\%, indicating grounding and navigation challenges when relying solely on visual input. In this setting, the best overall result comes from the hybrid setup Jedi-7B with GPT-4o (34.67\%), well above OpenAI CUA (20.98\%). These gaps highlight both the value of structured UI representations (A11y) and the advantage of specialized hybrid pipelines for UI grounding.

The Conversational category is where models perform best. With A11y inputs, Gemini Pro 2.5 reaches 74.07\%, while GPT-4o achieves 47.49\%. Under the Screenshots-only setting, OpenAI CUA performs best on MCQ questions. Multi-Dashboard remains the hardest category: even with A11y, the best accuracy is 15.22\% (Gemini Pro 2.5), reflecting persistent difficulties with cross-view reasoning and long-horizon planning. Overall, DASHBOARDQA exposes significant weaknesses in grounding, planning, and multi-view visual-mathematical reasoning.

\subsection{Qualitative Analysis}

We analyzed 20 randomly selected samples from the best-performing open-source and closed-source models—\textbf{Jedi-7B w/GPT-4o (Screenshot)} and \textbf{Gemini-Pro-2.5 (Screenshot + A11y Accessibility Tree)}, respectively. We traced the reasoning and execution steps taken by both models to identify underlying issues. The samples were drawn from cases where at least one of the models either failed to reach the correct answer or utilized the maximum number of reasoning steps. Our analysis revealed a number of common error patterns across both models.

\noindent \textbf{Plan Tracking Errors}: The models were observed to make various errors in tracking their initial reasoning plans during execution. While the initial plans were often well-structured, the models frequently failed to recognize that certain steps had already been completed. As a result, they redundantly reintroduced those steps into the intermediate plan, leading to repeated and unnecessary actions. For example, when the model is required to select three items sequentially from a dropdown list to update the dashboard view, it often fails to recall that the earlier selections have already been made. As a result, by the time it reaches the final step, it loops back and repeats the previous actions, creating a cycle that frequently hits the maximum limit of 25 steps. This is demonstrated in an example in (a) in figure \ref{fig:moreques_type}, where despite having all the details necessary from four views by step 9, the VLM has repeated looped over them repeatedly until the maximum number of steps was reached.

\noindent \textbf{Incorrect Information Retention}: One of the prominent issues affecting the models was their inability to retain the correct information or their tendency to hold on to incorrect details. This often led to flawed reasoning and ultimately incorrect answers. Sub figure (c) in figure \ref{fig:moreques_type} shows a case where the VLM misremembers the year it used for a preceding step.
In some cases, screenshots were captured too quickly, before the dashboard views had fully updated. Due to this delay, the models could end up retaining the wrong information as well.

\noindent \textbf{Incorrect Reasoning}: Models were also found to make blatant errors, or hallucinations, in their reasoning processes. Subfigure (b) demonstrates a clear visual reasoning error from a multi-line chart that ultimately led to an erroneous response.

Figure~\ref{fig:moreques_type} presents representative errors, while additional examples are provided in ~\ref{fig:extra_incorrect_examples}.

\subsection{Quantitative Analysis}
We further conducted a quantitative comparison of the reasoning behaviors exhibited by the strongest model under each setup: Gemini-Pro-2.5 (Screenshot + A11y setup) and Jedi-7B w/ GPT-4o (Screenshot setup). Specifically, we analyzed four key metrics: \emph{response lengths, step distributions, frequency of reaching the maximum step limit, and action type frequencies}. Figure~\ref{fig:steps-thoughts-comparison} illustrates the distributions of steps and reasoning thought lengths for both models. To avoid skewing the results, we excluded 11 extreme outliers in Gemini’s thought lengths (exceeding 10K characters).  For Jedi-7B, the reasoning traces are relatively short, with all actions restricted to simple \texttt{click} operations. Nevertheless, the model frequently fails to complete tasks within the allowed trajectory length, reaching the maximum 25-step limit in 56 cases. In contrast, Gemini exhibits more diverse interaction patterns, with actions distributed across three categories: \texttt{click} (2,920 instances), \texttt{moveTo} (3,796 instances), and \texttt{hotkey} (876 instances). Despite this richer action space, Gemini Pro 2.5 hits the 25-step ceiling in 77 cases.

\section{Conclusion}

We introduced \dashInteractqa{}, the first benchmark for question answering over interactive dashboards, aimed at advancing the capabilities of vision-language and GUI agents in realistic, exploratory data analysis tasks. Unlike prior benchmarks focused on static charts or utility-based GUI tasks, \dashInteractqa{} requires agents to plan and execute complex sequences of GUI interactions to answer diverse question types including factoid, multiple-choice, hypothetical, and conversational queries. Our experiments on this benchmark expose critical limitations in current models, with even the strongest agents experiencing substantial performance drops. Through extensive evaluation and analysis, we identify key challenges in grounding, planning, and visual reasoning, highlighting open problems in interactive multimodal understanding. We hope \dashInteractqa{} catalyzes future research toward more robust, grounded, and reasoning-capable agents for dynamic, real-world visual environments.

\section*{Limitations}

Even though there was a great effort taken to develop a holistic benchmark for evaluating VLMs capabilities, inherent limitations of the dataset must be noted. Firstly, this is a tableau-centric dataset. This limits the visual and topic diversity that would exist otherwise if dashboards from alternative visualization platforms such as Power BI and Zoho analytics were included.  In addition, the website URLS on tableau public were utilized as they allowed for accessibility tree access, allowing for smooth, no fault agentic navigation evaluation. This is not going to be available in desktop apps.\\ 
Future work on interactive dashboard question answering can focus more on essential GUI grounding tasks on the desktop app. \\

\section*{Ethical Considerations}

During the dashboard curation process authors were instructed to not collect controversial, divisive, or harmful dashboards, and to rather focus on more neutral and academic sources. This includes of sources including of United Nation subgroups and university institutions (i.e Oxford University).  All dashboards were taken from Tableau Public, where they are publicly available and permitted for use in developing AI models\footnote{\url{https://public.tableau.com/app/data-policy}}

All QA pairs were either fully authored by the authors or generated by vision-language models (VLMs) and subsequently reviewed and validated by the authors. Each author has expertise in data visualization comprehension tasks. The annotators were also informed that their annotations would be released with the dataset for research purposes. Additionally, AI-writing tools were employed to improve the clarity, structure, and overall writing quality of the paper.

\section*{Acknowledgement}
This work was supported by the Natural Sciences Engineering Research Council (NSERC) of Canada and Canada Foundation for Innovation (CFI). It also received support through a Google Cloud Platform (GCP) credits award from Google's PaliGemma Academic Program.

\bibliography{dashInteractQA}
\newpage
\appendix
\section{Appendices}

\subsection{Dataset Analysis}
\label{app:data_analysis}

\begin{table*}[t]
\centering
\caption{Counts of question creation methods through Human or VLMs.}
\label{tab:question_source_distribution}
\resizebox{\textwidth}{!}{%
\begin{tabular}{l|c|ccc|ccc}
\toprule
\multirow{2}{*}{\textbf{Question Category}} 
& \textbf{Human-made} 
& \multicolumn{3}{c|}{\textbf{Assisted}} 
& \multicolumn{3}{c}{\textbf{Model-Only}} \\
\cmidrule(lr){2-2}\cmidrule(lr){3-5}\cmidrule(lr){6-8}
&  
& \textbf{GPT4o} 
& \textbf{Claude} 
& \textbf{Gemini} 
& \textbf{GPT4o} 
& \textbf{Claude} 
& \textbf{Gemini} \\
\midrule
MCQ             & 39 & 15 & 10 & 7  & 5 & 1 & 3 \\
Factoid         & 34 & 11 & 0  & 6  & 3 & 0 & 13 \\
Hypothetical    & 28 & 19 & 8  & 6  & 1 & 0 & 0 \\
Multi-Dashboard & 18 & 8  & 8  & 10 & 2 & 0 & 0 \\
Conversational  & 9  & 7  & 6  & 6  & 0 & 0 & 0 \\
\bottomrule
\end{tabular}
}
\end{table*}
\subsubsection{Dashboard topical and source diversity}
\label{app:visual_diversity}
\begin{figure*}[t]
    \includegraphics[width=\textwidth]{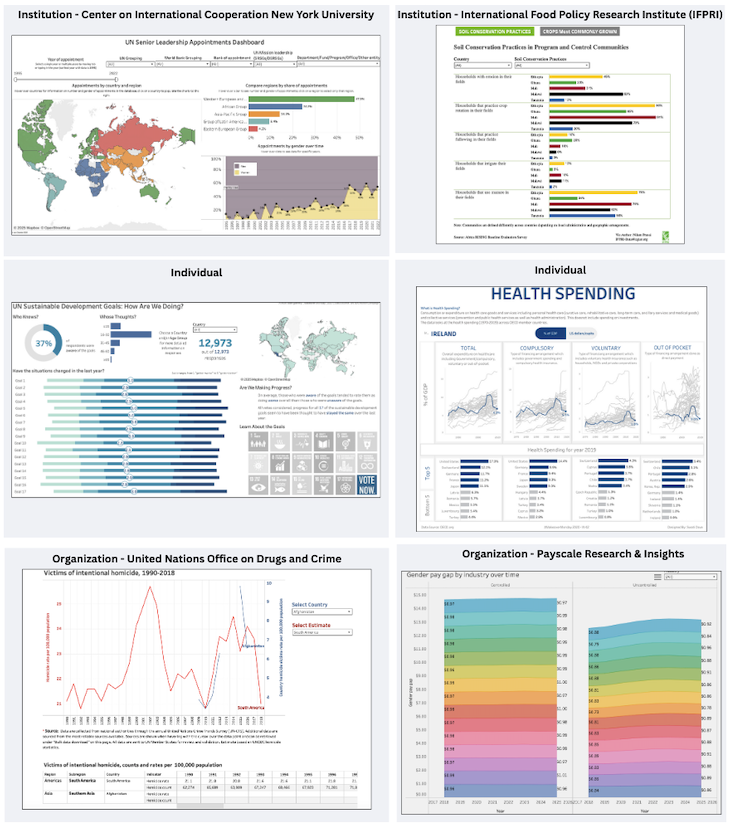}
    \caption{Screenshot examples of tableau dashboards collected from various institutions,organizations,and individuals}
    \label{fig:src_charts}
\end{figure*}

\begin{figure*}[t]
    \includegraphics[width=\textwidth]{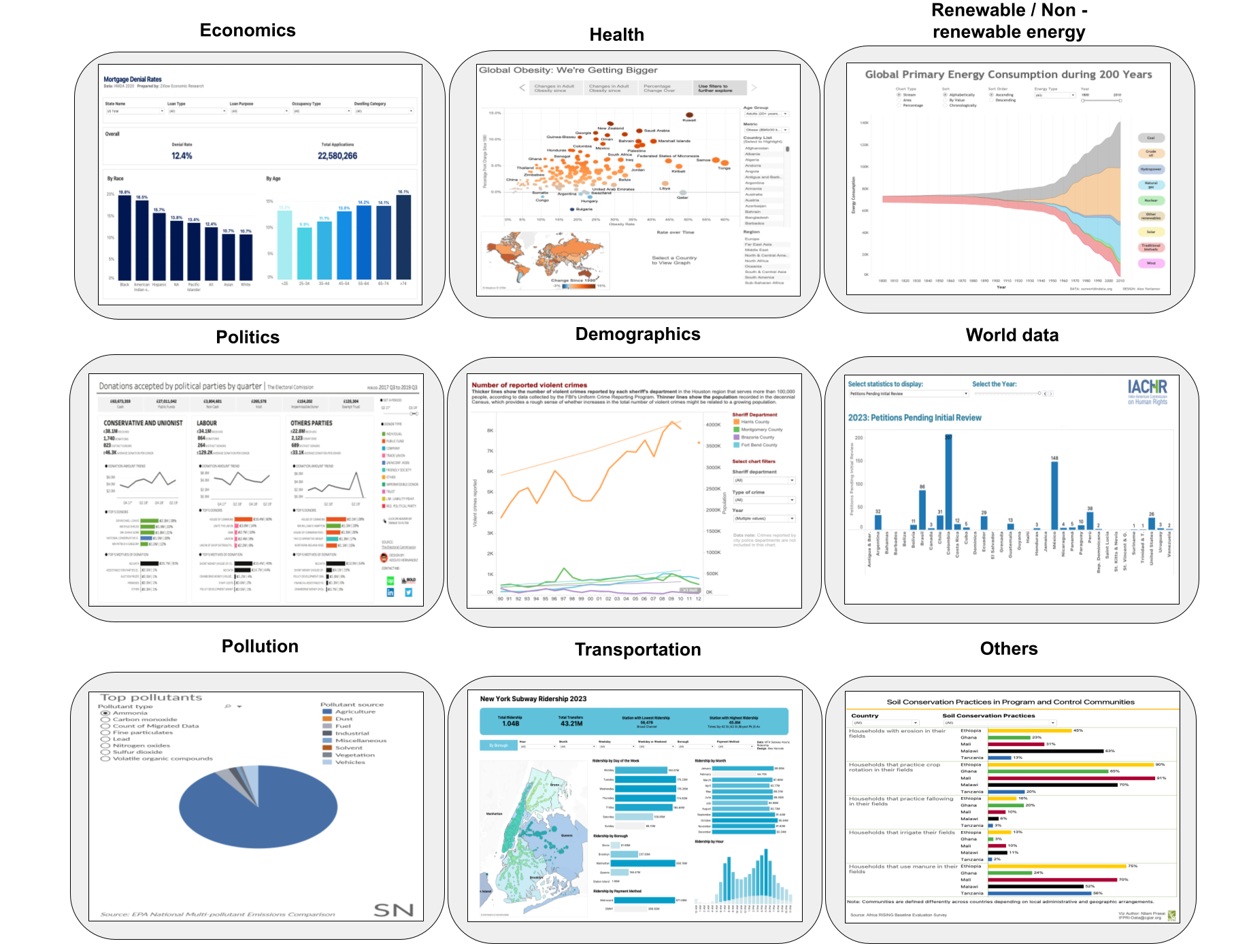}
    \caption{Sample dashboards chosen for all of the topics in DashboardQA}
    \label{fig:topic_charts}
\end{figure*}

Figure \ref{fig:topic_charts} shows example dashboards from all of the topics covered, and figure \ref{fig:src_charts} shows example dashboards from different source types. 

\subsection{Question source makeup}
\label{app:Question_diversity}
Table \ref{tab:question_source_distribution} depicts the distribution of the sources for the question answer pairs generated.

\begin{figure*}[t]
    \includegraphics[width=\textwidth,keepaspectratio,
                     trim=1.2cm 1.0cm 1.2cm 1.0cm,clip]{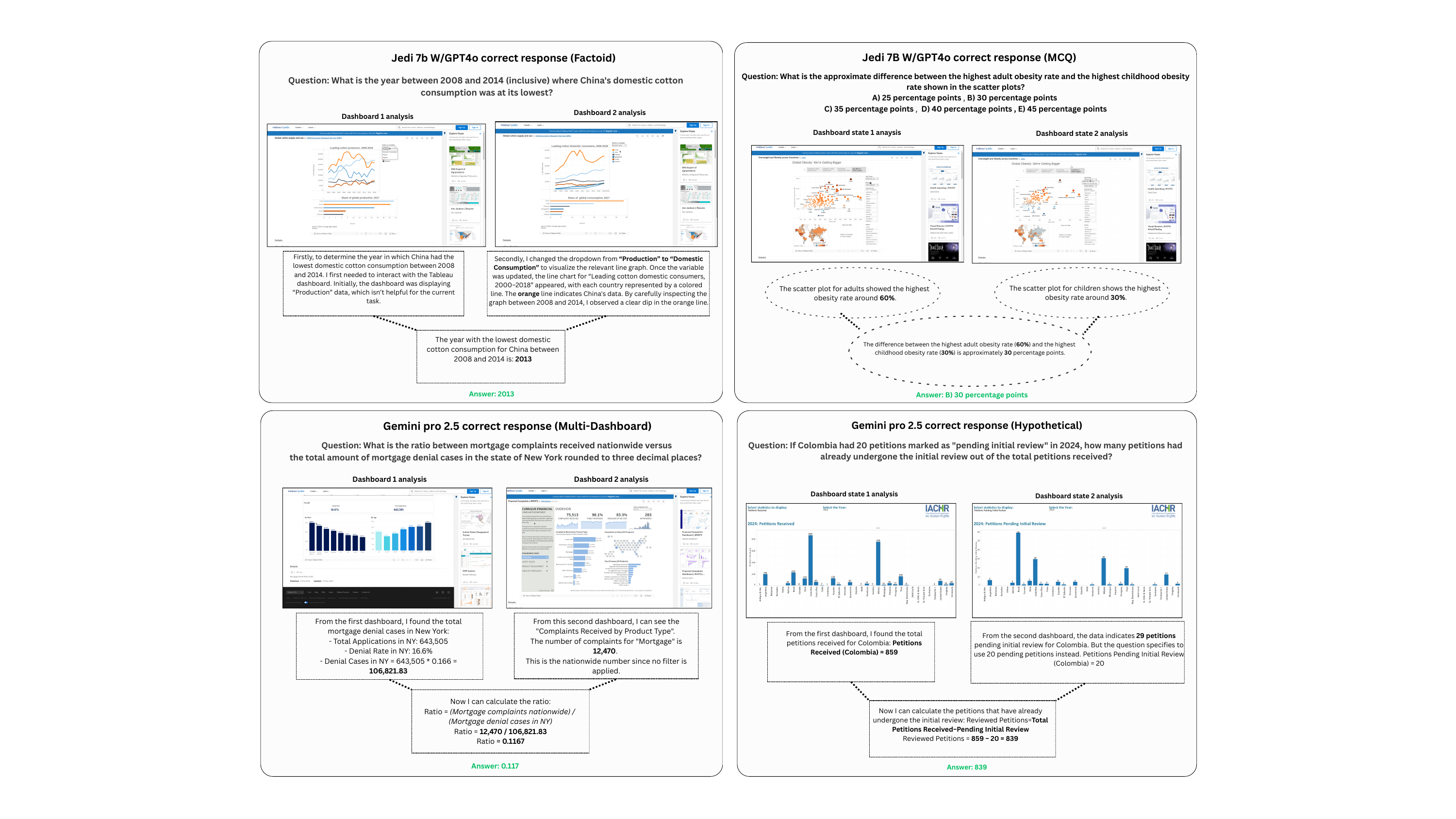}
    \caption{Examples of QA pairs that have received correct responses, and their corresponding dashboard and VLM.}
    \label{fig:llm_gen}
\end{figure*}

\begin{figure*}[t]
    \includegraphics[width=\linewidth]{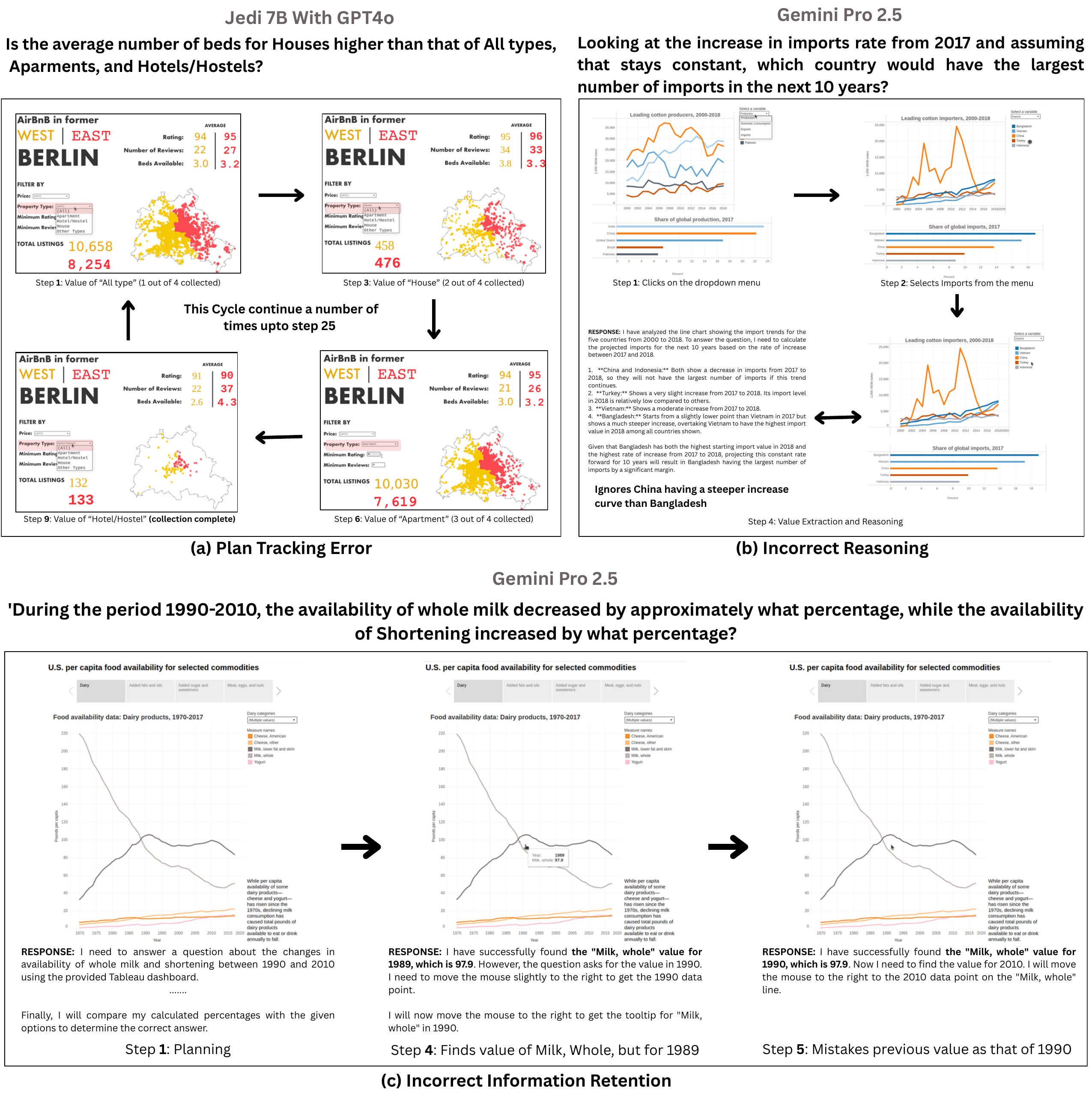}
    \caption{Examples types of errors seen}
    \label{fig:moreques_type}
\end{figure*}

\begin{figure*}[t]
    \includegraphics[width=\textwidth]{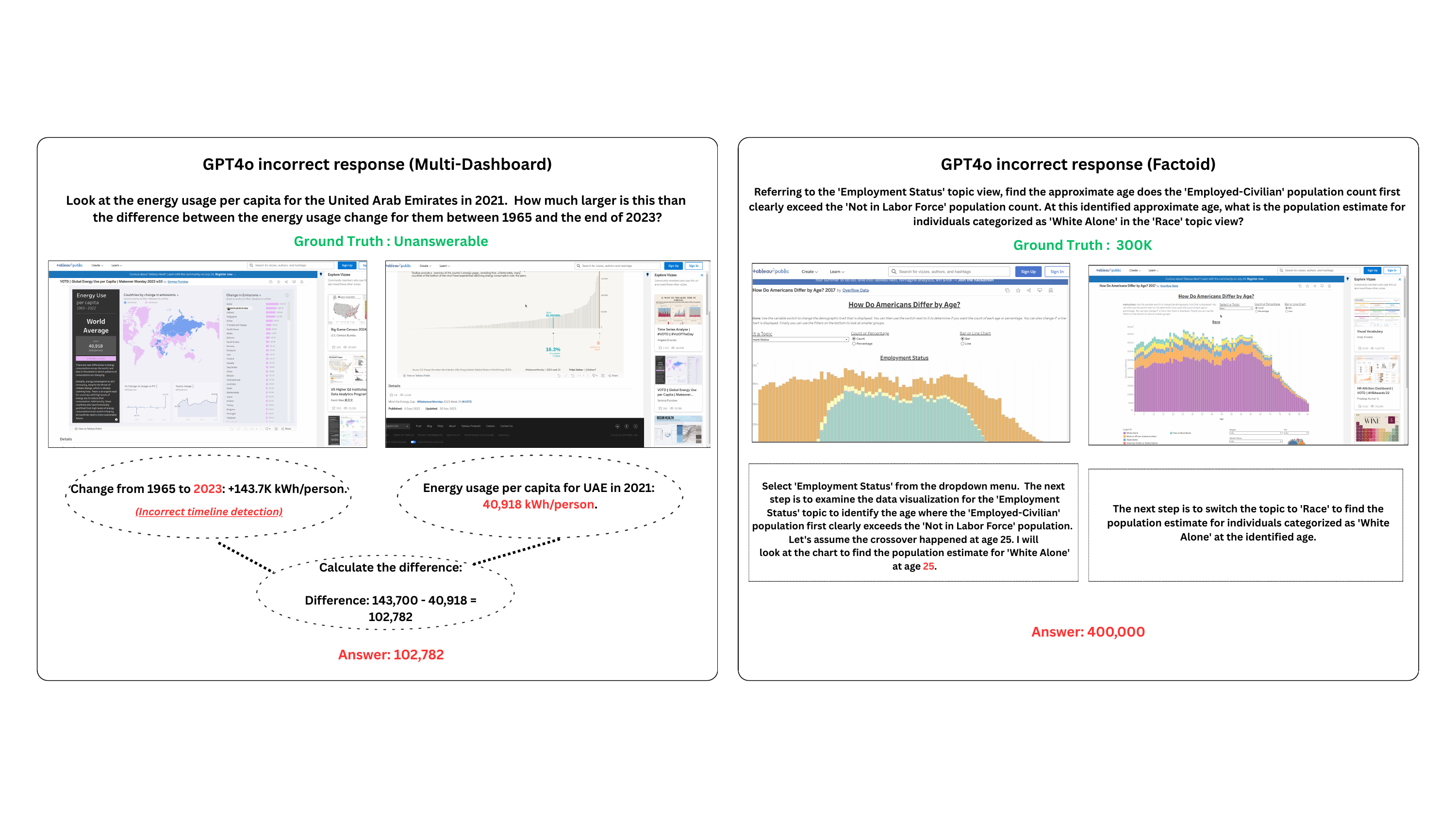}
    \caption{Additional examples of errors, as done by GPT4o.}
    \label{fig:extra_incorrect_examples}
\end{figure*}

\subsection{Prompts for QA pair generation}
\label{app:prompts-evaluation}

For reproducibility, we provide the exact prompts used to generate QA pairs in Tables\ref{tab:prompt_templates_direct} and \ref{tab:prompt_templates_cot}.

\begin{table*}[htbp]
  \centering
  \caption{Prompt Templates for Factoid, MCQ, and Hypothetical question answer generation. These represent the single turn, single dashboard question types.}
  \label{tab:prompt_templates_direct}
  \rowcolors{2}{gray!10}{white}
  \begin{tabularx}{\textwidth}{>{\bfseries}l X}
    \toprule
    Category & Prompt Template \\ 
    \midrule
    Factoid & 
    \begin{minipage}[t]{\linewidth}
      \footnotesize	
      I am giving you multiple snapshots/views of multiple interactive dashboards, including of their respective default views, options available for navigation through the dashboard’s navigation tools (i.e Dropdown menus, radio buttons, scrollers, etc.), and more views from when particular options are selected. Since not every possible snapshot / view of each dashboard is given, the selection options available and the example views for each dashboard should together help you guess how the views for the rest of the options would be structurally (i.e types and content of the charts or textual data available) for that particular dashboard.
\\
Using all of these, please generate 5 diverse and challenging factoid reasoning questions that involve arithmetic and logical reasoning. They must have multiple operators (i.e sum, ratio, etc.) in the question. Additionally, make sure the answers required are succinct, where they are either a numerical value, True or False, Yes or No, unanswerable or simply a label taken from the dashboard itself. Each question must require 3 to 5 views (Cumulative) of the dashboards to answer. Just the questions are to be given now from your end, in the next prompt the required views will be given for you to answer.
\\
Here are some sample questions to give some inspiration, but make sure to create the questions with lexical and semantic differences.
Samples : \textless samples \textgreater
    \end{minipage} \\
    \midrule
    Multi Choice & 
    \begin{minipage}[t]{\linewidth}
      \footnotesize	
      I am giving you multiple snapshots/views of multiple interactive dashboards, including of their respective default views, options available for navigation through the dashboard’s navigation tools (i.e Dropdown menus, radio buttons, scrollers, etc.), and more views from when particular options are selected. Since not every possible snapshot / view of each dashboard is given, the selection options available and the example views for each dashboard should together help you guess how the views for the rest of the options would be structurally (i.e types and content of the charts or textual data available) for that particular dashboard.
\\
Using all of these, please generate 5 diverse and challenging MCQ reasoning questions that involve arithmetic and logical reasoning. There may be 4-5 answer options available, but only one of them is correct. They must have multiple operators (i.e sum, ratio, etc.) in the question. Additionally, make sure the answers required are succinct, where they are either a numerical value, True or False, Yes or No, unanswerable or simply a label taken from the dashboard itself. Each question must require 2 to 5 views of the dashboard to answer. Just the questions are to be given now from your end, in the next prompt the required views will be given for you to answer.
Remember to not be biased towards just the views given, and to reason as to how the others would look like as to help your questions. 
\\
Here are some sample questions to give some inspiration, but make sure to create the questions with lexical and semantic differences.
Samples : \textless samples \textgreater
    \end{minipage} \\
    \midrule
    Hypothetical & 
    \begin{minipage}[t]{\linewidth}
      \footnotesize	
     I am giving you multiple snapshots/views of multiple interactive dashboards, including of their respective default views, options available for navigation through the dashboard’s navigation tools (i.e Dropdown menus, radio buttons, scrollers, etc.), and more views from when particular options are selected. Since not every possible snapshot / view of each dashboard is given, the selection options available and the example views for each dashboard should together help you guess how the views for the rest of the options would be structurally (i.e types and content of the charts or textual data available) for that particular dashboard.
\\
Using all of these, please generate 5 diverse and challenging hypothetical questions that involve arithmetic and logical reasoning. This may involve looking into future trends, extrapolation tasks, or looking at counterfactuals for example. They must have multiple operators (i.e sum, ratio, etc.) in the question.  Additionally, make sure the answers required are succinct, where they are either a numerical value, True or False, Yes or No, unanswerable or simply a label taken from the dashboard itself. Each question must require from 2 to 5 views of the dashboard to answer. Just the questions are to be given now from your end, in the next prompt the required views will be given for you to answer.
\\
Here are some sample questions to give some inspiration, but make sure to create the questions with lexical and semantic differences.
Samples : \textless samples \textgreater
    \end{minipage} \\
    \midrule \\
    \bottomrule
  \end{tabularx}
\end{table*}

\begin{table*}[htbp]
  \centering
  \caption{Prompt Templates for Multi-Dashboard and conversational question answer generation. These represent the multi-turn or multi-dashboard question types. }
  \label{tab:prompt_templates_cot}
  \rowcolors{2}{gray!10}{white}
  \begin{tabularx}{\textwidth}{>{\bfseries}l X}
    \toprule
    Category & Prompt Template \\ 
    \midrule
    Multi-Dashboard &
    \begin{minipage}[t]{\linewidth}
      \small
      I am giving you multiple snapshots/views respectively for multiple interactive dashboards, including of their respective default views, options available for navigation through the dashboard’s navigation tools (i.e Dropdown menus, radio buttons, scrollers, etc.), and more views from when particular options are selected. Since not every possible snapshot / view of each dashboard is given, the selection options available and the example views for each dashboard should together help you guess how the views for the rest of the options would be structurally (i.e types and content of the charts or textual data available) for that particular dashboard.
\\
Using all of these, please generate 5 diverse and high level questions surrounding all of the dashboards given that involve arithmetic and logical reasoning. This could include having multiple operators (i.e sum, ratio, etc.) in the question. Additionally, make sure the answers required are succinct, where they are either a numerical value, True or False, Yes or No, unanswerable or simply a label taken from the dashboard itself. Each question must require from 2 to 5 views of the dashboard to answer, and a majority of the follow up questions should require utilizing one or more of the previous questions to help. Just the questions are to be given now from your end, in the next prompt the required views will be given for you to answer.
\\
Here are some sample questions to give some inspiration, but make sure to create the questions with lexical and semantic differences.
Samples : \textless samples \textgreater
    \end{minipage} \\
    \midrule
    Conversational &
    \begin{minipage}[t]{\linewidth}
      \small
          I am giving you multiple snapshots/views of an interactive dashboard, including the default view, the options available for navigation through the dashboard’s navigation tools (i.e Dropdown menus, radio buttons, scrollers, etc.), and more views from when particular options are selected. Since not every possible snapshot / view are going to be given, the selection options available and the example views should together help you guess how the views for the rest of the options would be structurally (i.e types and content of the charts or textual data available for that particular dashboard).
\\
Using all of these, please generate a conversation of 4-7 questions that analyzes the contents of the dashboard and requires multiple operations (i.e sum, ratio etc.). With a focus on arithmetic and logical reasoning, make sure that a majority of the questions follow up on each other. Use ambiguous pronouns in the latter questions when making references to previous questions. Additionally, make sure the answers required are succinct, where they are either a numerical value, True or False, Yes or No, or simply a label taken from the dashboard itself. The conversation must require from 2 to 5 views of the conversation to answer, where each individual question can either require just one to three views to answer respectively. Just the questions are to be given now from your end, in the next prompt the required views will be given for you to answer.
\\
Here are some sample conversations to give some inspiration, but make sure to create the conversations with different lexical and semantic differences. The respective dashboard images are not given for these samples, they are just to get an idea (Not purely representative) of the style of conversations.

Samples : \textless samples \textgreater
    \end{minipage} \\
    \midrule
    \bottomrule
  \end{tabularx}
\end{table*}

\subsection{Prompts for Models Evaluatxion}
\label{app:prompts-evaluation}

For reproducibility, we provide the prompts used for the models evaluation, adapted from the ChartQAPro benchmark ~\citep{masry2025chartqaprodiversechallengingbenchmark} in Table \ref{tab:prompt_templates}. 

\begin{table*}[htbp]
  \centering
  \caption{Prompt Templates for Each Question Category in the Direct setup.}
  \label{tab:prompt_templates}
  \rowcolors{2}{gray!10}{white}
  \begin{tabularx}{\textwidth}{>{\bfseries}l X}
    \toprule
    Category & Prompt Template \\ 
    \midrule
    Factoid & 
    \begin{minipage}[t]{\linewidth}
      \small

      You are given a factoid question about an interactive Tableau dashboard that you need to navigate to answer the question.
    You need to think step-by-step, but your final answer should be a single word, number, or phrase. Do not generate units. But if numerical units such as million, m, billion, B, or K are required, use the exact notation shown in the dashboard.
    Remember to navigate the dashboard, think step-by-step, and put the final answer between these brackets <answer> </answer>

    Question:  \textless question\textgreater

    \end{minipage} \\
    \midrule
    Multi Choice & 
    \begin{minipage}[t]{\linewidth}
      \small
      You are given a question about an interactive Tableau dashboard along with different possible answers. You need to navigate the dashboard to select the correct answer from them.
        You need to think step-by-step, but your final answer should be one of the options letters only (without any additional text). 
        Remember to navigate the dashboard, think step-by-step, and put the final answer between these brackets <answer> </answer>.

      Question: \textless question\textgreater
    \end{minipage} \\
    \midrule
    Hypothetical & 
    \begin{minipage}[t]{\linewidth}
      \small
      You are given a hypothetical question about an interactive Tableau dashboard that you need to navigate to answer the question.
        You need to think step-by-step, but your final answer should be a single word, number, or phrase. 
        Do not generate units. But if numerical units such as million, m, billion, B, or K are required, use the exact notation shown in the dashboard.
        Remember to navigate the dashboard, think step-by-step, and put the final answer between these brackets <answer> </answer>
        
      Question: \textless question\textgreater
    \end{minipage} \\
    \midrule
    Multi Dashboards & 
    \begin{minipage}[t]{\linewidth}
      \small
      You are given a question about two interactive Tableau dashboards that are open in two tabs in the browser. You need to navigate them to answer the question.
        You need to think step-by-step, but your final answer should be a single word, number, or phrase. 
        Do not generate units. But if numerical units such as million, m, billion, B, or K are required, use the exact notation shown in the dashboard.
        Remember to navigate the dashboard, think step-by-step, and put the final answer between these brackets <answer> </answer>
      Question: \textless question\textgreater
    \end{minipage} \\
    \midrule
    Conversational & 
    \begin{minipage}[t]{\linewidth}
      \small
      You are given a multi-turn conversation, and your job is to answer the final question based on the conversation history and the information in the provided interactive Tableau dashboard that you need to navigate.
        You need to think step-by-step, but your final answer should be a single word, number, or phrase. 
        Do not generate units. But if numerical units such as million, m, billion, B, or K are required, use the exact notation shown in the dashboard.
        Remember to navigate the dashboard, think step-by-step, and put the final answer between these brackets <answer> </answer>
        
      Question: \textless question with conversation history\textgreater
    \end{minipage} \\
    \bottomrule
  \end{tabularx}
\end{table*}

\subsection{Quantitative Analysis}
\label{app:quant-analysis}
\begin{figure*}[t]
    \centering
    \begin{subfigure}[b]{0.48\textwidth}
        \centering
        \includegraphics[width=\linewidth]{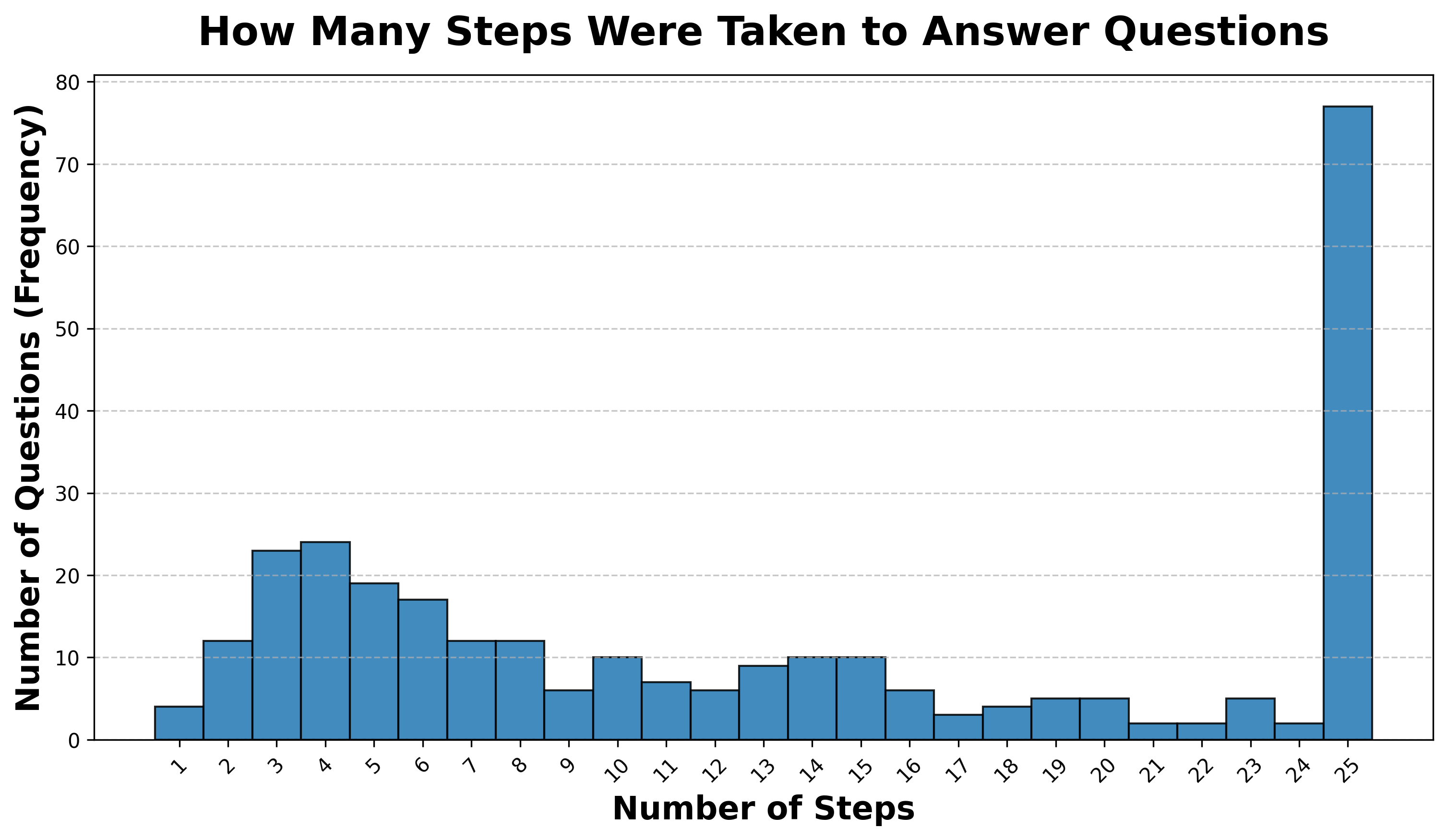}
        \caption{Distribution of steps taken by \textbf{Gemini-Pro-2.5}.}
    \end{subfigure}
    \hfill
    \begin{subfigure}[b]{0.48\textwidth}
        \centering
        \includegraphics[width=\linewidth]{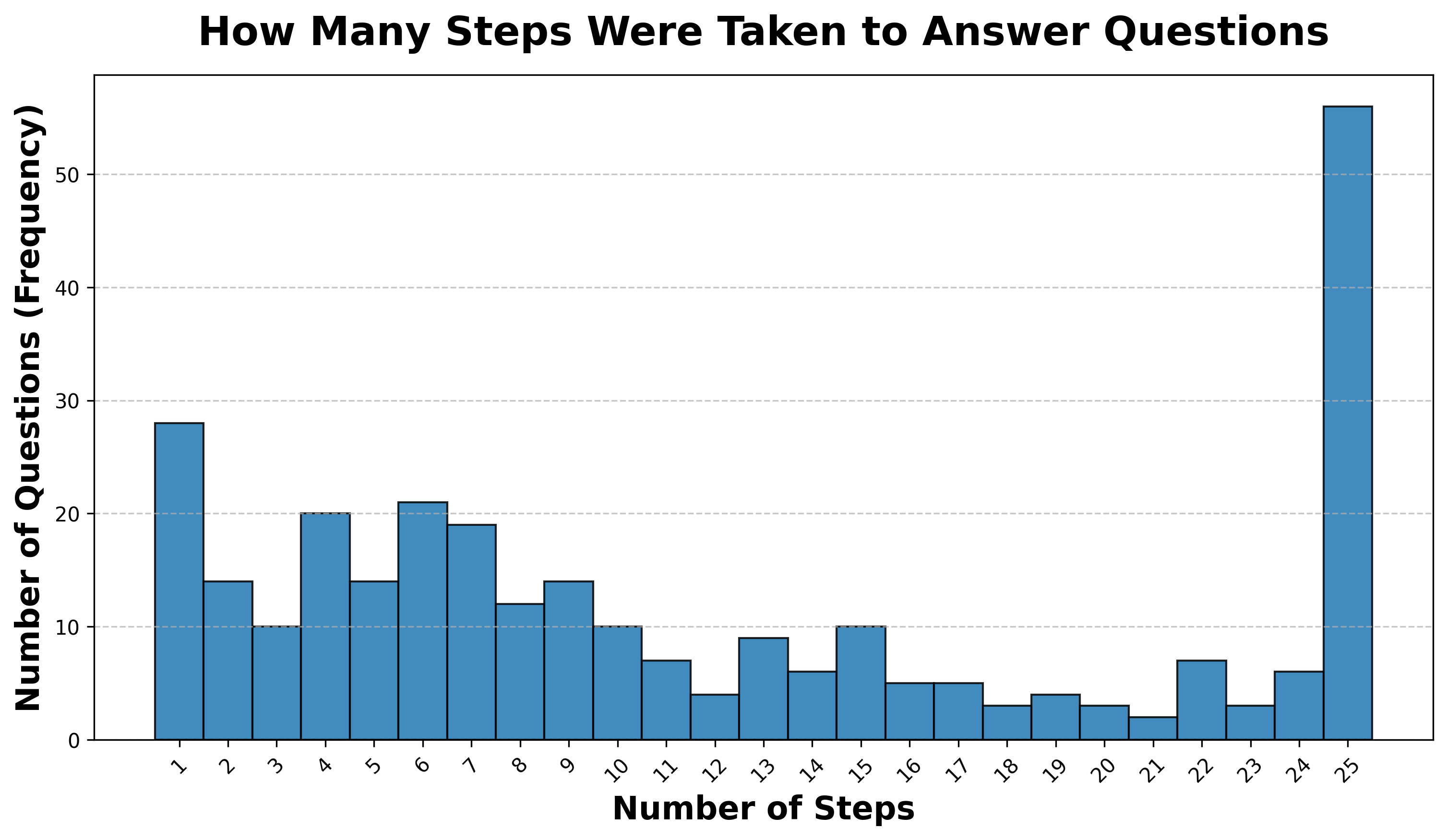}
        \caption{Distribution of steps taken by \textbf{Jedi-7B w/ GPT-4o}.}
    \end{subfigure}

    \begin{subfigure}[b]{0.48\textwidth}
        \centering
        \vspace{1em}
        \includegraphics[width=\linewidth]{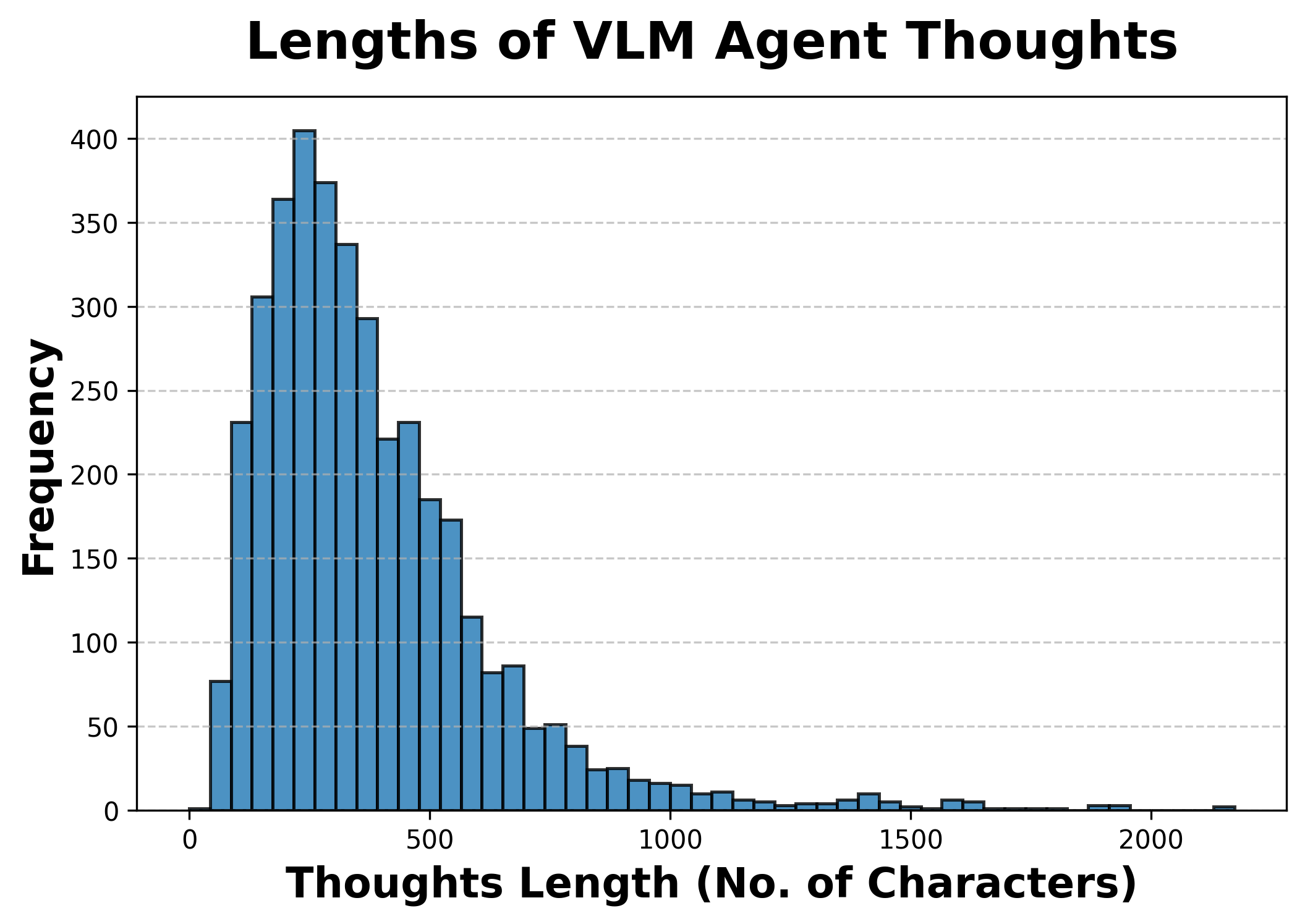}
        \caption{Lengths of reasoning thoughts generated by \textbf{Gemini}.}
    \end{subfigure}
    \hfill
    \begin{subfigure}[b]{0.48\textwidth}
        \centering
        \vspace{1em}\vspace{1em}
        \includegraphics[width=\linewidth]{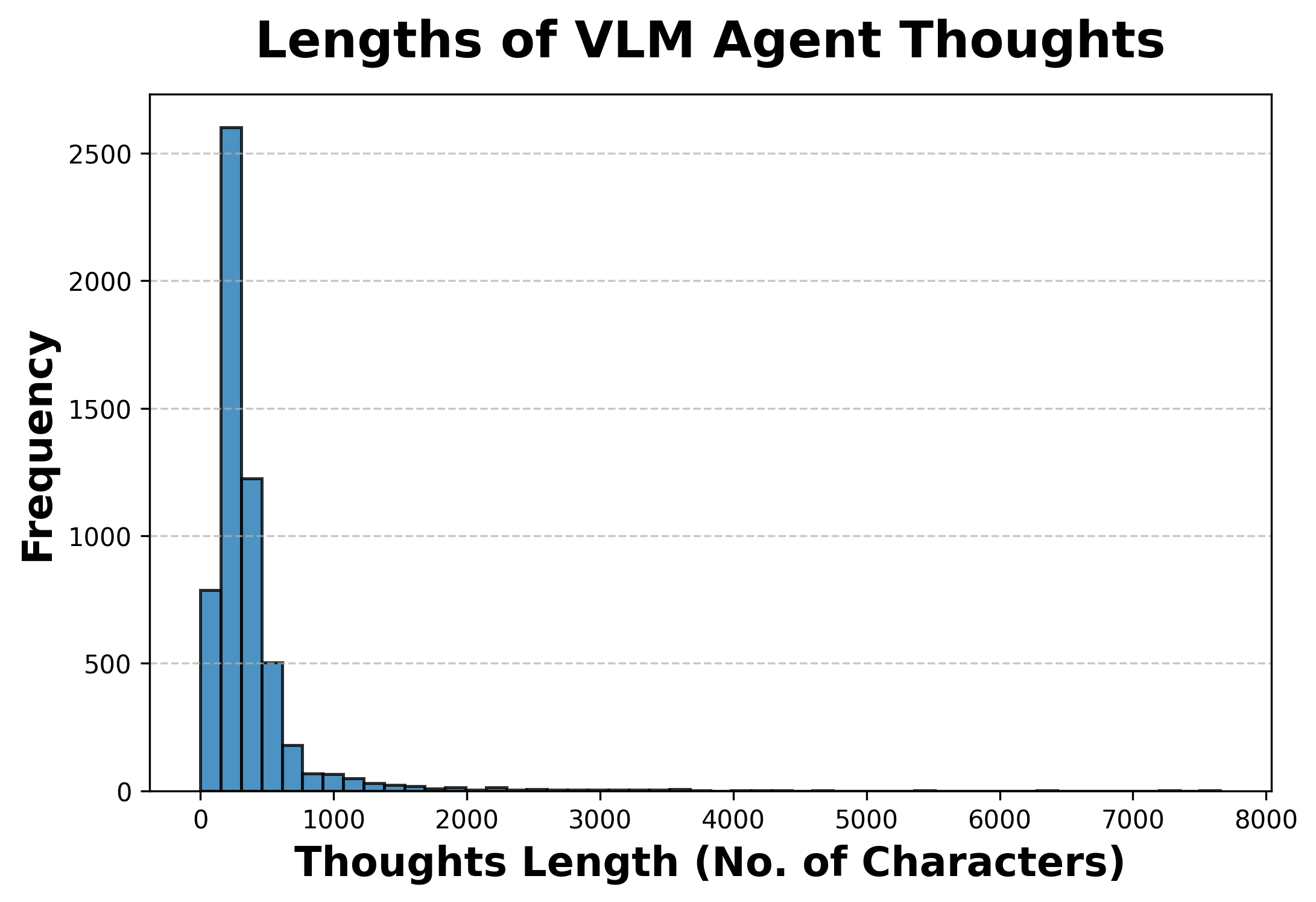}
        \caption{Lengths of reasoning thoughts generated by \textbf{Jedi-7B}.}
    \end{subfigure}

    \caption{Comparison of reasoning behaviors between Gemini-Pro-2.5 (left) and Jedi-7B w/ GPT-4o (right) on the \dashInteractqa{} benchmark. 
    The plots illustrate (a, b) the distribution of the number of interaction steps taken to answer questions and (c, d) the lengths of intermediate reasoning thoughts.}
    \label{fig:steps-thoughts-comparison}
\end{figure*}

Figure \ref{fig:steps-thoughts-comparison} shows the distribution of steps and thought lengths for both Gemini Pro 2.5 and Jedi 7B w/GPT4o. For clarity, we excluded 11 extreme outliers in Gemini’s thought lengths (exceeding 10K characters) to avoid skewing the histograms.

\end{document}